\documentclass{article} %
\usepackage{iclr2019_conference,times}

\usepackage{amsmath,amsfonts,bm}

\def\eqref#1{equation~\ref{#1}}

\def\floor#1{\lfloor #1 \rfloor}
\def\1{\bm{1}}

\DeclareMathAlphabet{\mathsfit}{\encodingdefault}{\sfdefault}{m}{sl}
\SetMathAlphabet{\mathsfit}{bold}{\encodingdefault}{\sfdefault}{bx}{n}

\usepackage{color,xcolor}
\usepackage{epsfig}
\usepackage{graphicx}
\usepackage{subfiles}

\usepackage{adjustbox}
\usepackage{array}
\usepackage{booktabs}
\usepackage{colortbl}
\usepackage{float,wrapfig}
\usepackage{hhline}
\usepackage{multirow}
\usepackage{subcaption} %
\usepackage[labelfont={bf},labelsep={period},font={small}]{caption}

\let\llncssubparagraph\subparagraph
\let\subparagraph\paragraph
\usepackage[compact]{titlesec}
\let\subparagraph\llncssubparagraph

\usepackage{amsmath,amsfonts,amssymb}
\usepackage{bm}
\usepackage{nicefrac}
\usepackage{microtype}

\usepackage{changepage}
\usepackage{extramarks}
\usepackage{fancyhdr}
\usepackage{lastpage}
\usepackage{setspace}
\usepackage{soul}
\usepackage{xspace}

\usepackage{url}
\usepackage{hyperref}

\usepackage{algorithm}
\usepackage{algpseudocode}
\usepackage{enumerate}

\newcolumntype{L}[1]{>{\raggedright\let\newline\\\arraybackslash\hspace{0pt}}m{#1}}
\newcolumntype{C}[1]{>{\centering\let\newline\\\arraybackslash\hspace{0pt}}m{#1}}
\newcolumntype{R}[1]{>{\raggedleft\let\newline\\\arraybackslash\hspace{0pt}}m{#1}}

\newcommand{\ignore}[1]{}

\makeatletter
\DeclareRobustCommand\onedot{\futurelet\@let@token\@onedot}
\def\@onedot{\ifx\@let@token.\else.\null\fi\xspace}

\def\eg{\emph{e.g}\onedot} 
\def\ie{\emph{i.e}\onedot} 
 
\def\etc{\emph{etc}\onedot} 
\def\wrt{w.r.t\onedot} 

\makeatother

\definecolor{MyDarkBlue}{rgb}{0,0.08,1}
\definecolor{MyDarkGreen}{rgb}{0.02,0.6,0.02}
\definecolor{MyDarkRed}{rgb}{0.8,0.02,0.02}
\definecolor{MyDarkOrange}{rgb}{0.40,0.2,0.02}
\definecolor{MyPurple}{RGB}{111,0,255}
\definecolor{MyRed}{rgb}{1.0,0.0,0.0}
\definecolor{MyGold}{rgb}{0.75,0.6,0.12}
\definecolor{MyDarkgray}{rgb}{0.66, 0.66, 0.66}

\algnewcommand{\LeftComment}[1]{\Statex \(\triangleright\) #1}

\newcommand{\methodnameverbose}{Defensive Quantization (DQ)\xspace}
\newcommand{\methodname}{Defensive Quantization\xspace}

\title{Defensive Quantization: \\ When Efficiency Meets Robustness}

\author{Ji Lin \\
MIT\\
\texttt{jilin@mit.edu} \\
\And
Chuang Gan \\
MIT-IBM Watson AI Lab \\
\texttt{ganchuang@csail.mit.edu} \\
\And
Song Han \\
MIT \\
\texttt{songhan@mit.edu}
}

\iclrfinalcopy %
\begin{document}

\maketitle

\begin{abstract}
 Neural network quantization is becoming an industry standard to efficiently deploy deep learning models on hardware platforms, such as CPU, GPU, TPU, and FPGAs. However, we observe that the conventional quantization approaches are vulnerable to adversarial attacks. This paper aims to raise people's awareness about the security of the quantized models, and we designed a novel quantization methodology to jointly optimize the efficiency and robustness of deep learning models. We first conduct an empirical study to show that vanilla quantization suffers more from adversarial attacks. We observe that the inferior robustness comes from the \emph{error amplification effect}, where the quantization operation further enlarges the distance caused by amplified noise. Then we propose a novel \methodnameverbose method by controlling the Lipschitz constant of the network during quantization, such that the magnitude of the adversarial noise remains non-expansive during inference. Extensive experiments on CIFAR-10 and SVHN datasets demonstrate that our new quantization method can defend neural networks against adversarial examples, and even achieves superior robustness than their full-precision counterparts, while maintaining the same hardware efficiency as vanilla quantization approaches. As a by-product, DQ can also improve the accuracy of quantized models without adversarial attack. 

\end{abstract}

\section{Introduction}

\begin{figure}[t]
\centering
\begin{subfigure}[b]{0.45\textwidth}
	\includegraphics[width=\textwidth]{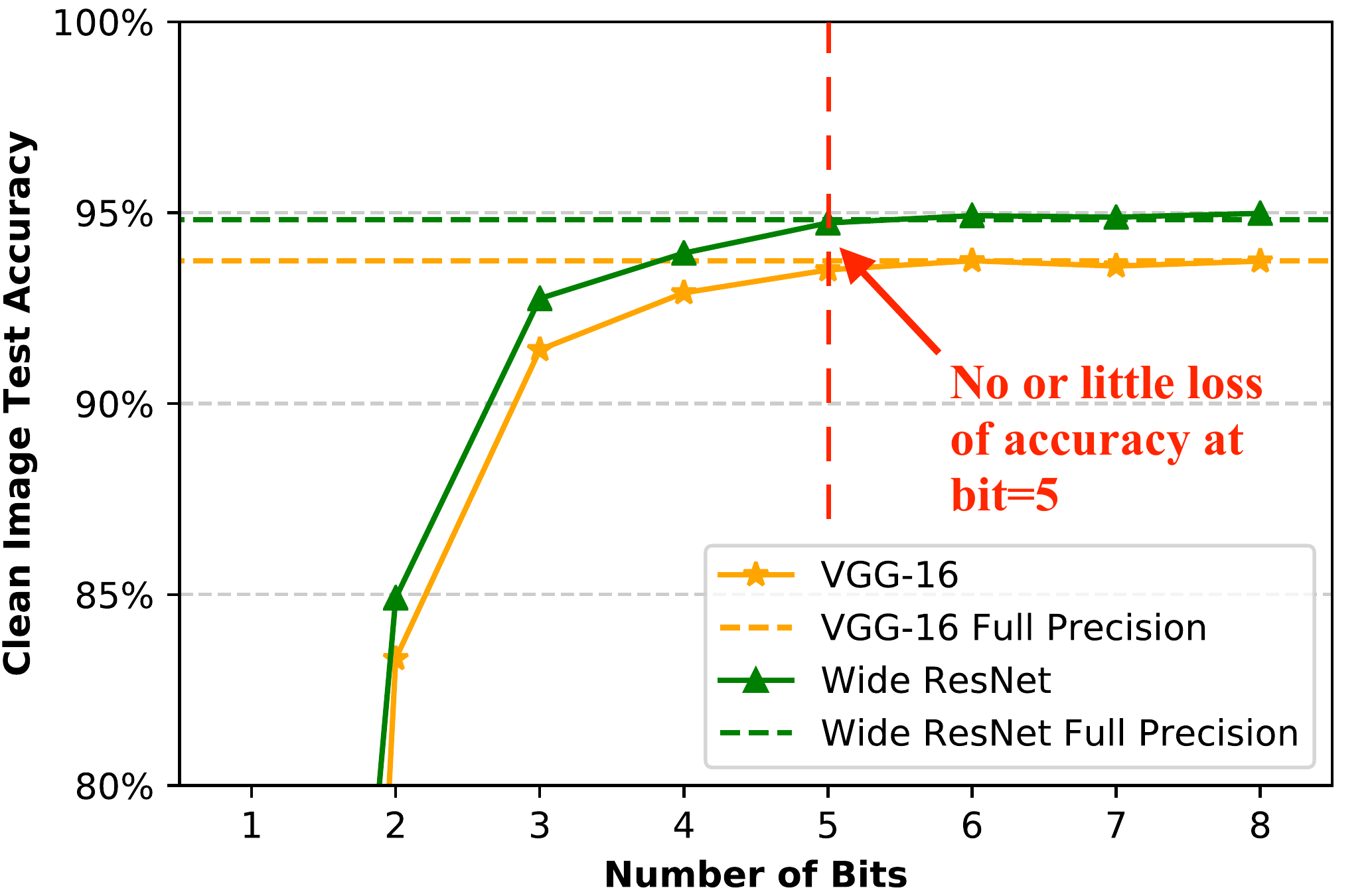}
	\caption{Quantization preserves the accuracy till 4-5 bits on clean image.}
	\label{fig:clean_acc_bit_curve}
\end{subfigure}
~~~~
\begin{subfigure}[b]{0.45\textwidth}
	\includegraphics[width=\textwidth]{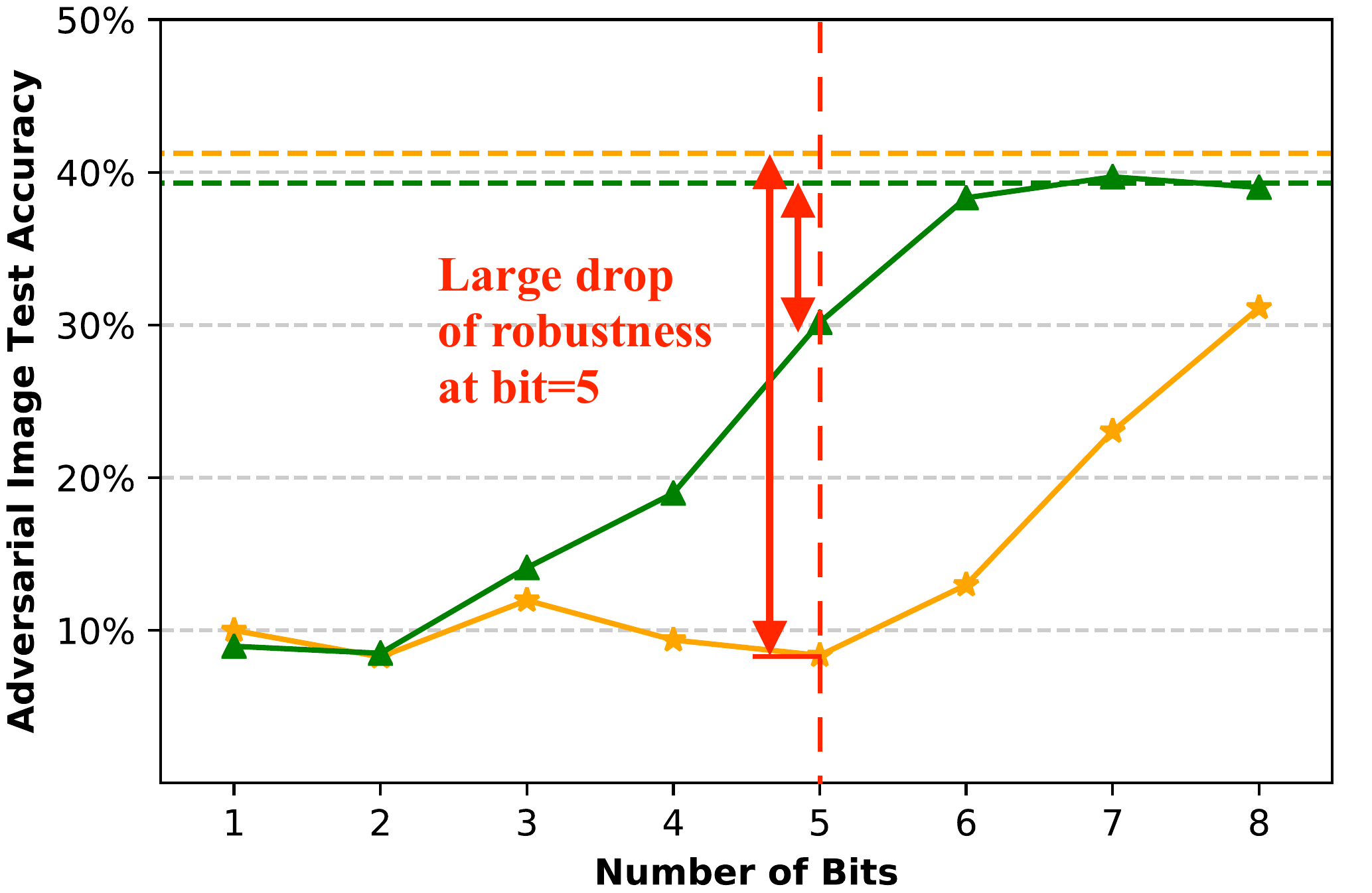}
	\caption{Quantization no longer preserves the accuracy under adversarial attack (same legend as left).}
	\label{fig:adv_acc_bit_curve}
\end{subfigure}
\caption{Quantized neural network are more vulnerable to adversarial attack.
Quantized models have no loss of accuracy on clean image ($\geq$5 bits), but have significant loss of accuracy under adversarial attack compared to full precision models.  Setup: VGG-16 and Wide ResNet on the test set of CIFAR-10 with FGSM attack.
}
\label{fig:acc_bit_curve}
\end{figure}

Neural network quantization~\citep{han2015deep, zhu2016trained, jacob2017quantization} is a widely used technique to reduce the computation and memory costs of neural networks, facilitating efficient deployment. It has become an industry standard for deep learning hardware.
However, we find that the widely used vanilla quantization approaches suffer from unexpected issues --- the quantized model is more vulnerable to adversarial attacks (Figure~\ref{fig:acc_bit_curve}). Adversarial attack is consist of subtle perturbations on the input images that causes the deep learning models to give incorrect labels~\citep{szegedy2013intriguing, goodfellow6572explaining}. Such perturbations are hardly detectable by human eyes but can easily fool neural networks. Since quantized neural networks are widely deployed in many safety-critical scenarios, \eg, autonomous driving~\citep{amodei2016concrete}, the potential security risks cannot be neglected. The efficiency and latency in such applications are also important, so we need to jointly optimize them.

The fact that quantization leads to inferior adversarial robustness is counter intuitive, as small perturbations should be denoised with low-bit representations. Recent work~\citep{xu2017feature} also demonstrates that quantization on input image space , \ie
color bit depth reduction, is quite effective to defend adversarial examples. A natural question then rises, why the quantization operator is yet effective when applied to intermediate DNN layers? We analyze that such issue is caused by the error amplification effect of adversarial perturbation~\citep{liao2018defense} --- although the magnitude of perturbation on the image is small, it is amplified significantly when passing through deep neural network (see Figure~\ref{fig:vgg16_layer_diff}). The deeper the layers are, the more significant such side effect is. Such amplification pushes values into a different quantization bucket, which is undesirable. We conducted empirical experiments to analyze how quantization influences the activation error between clean and adversarial samples (Figure~\ref{fig:dist_vs_epsilon}): when the magnitude of the noise is small, activation quantization is capable of reducing the errors by eliminating small perturbations; However, when the magnitude of perturbation is larger than certain threshold, quantization instead amplify the errors, which causes the quantized model  to make mistakes. We argue that this is the main reason causing the inferior robustness of the quantized models.

In this paper, we propose \methodnameverbose that not only fixes the robustness issue of quantized models, but also turns activation quantization into a defense method that further boosts adversarial robustness. We are inspired by the success of image quantization in improving robustness. 
Intuitively, it will be possible to defend the attacks with quantization operations if we can keep the magnitude of the perturbation small. However, due to the error amplification effect of gradient based adversarial samples, it is non-trivial to keep the noise at a small scale during inference. Recent works~\citep{cisse2017parseval, qian2018l2} have attempted to make the network \emph{non-expansive} by controlling the Lipschitz constant of the network to be smaller than 1, which has smaller variation change in its output than its input. In such case, the input noise will not propagate through the intermediate layers and impact the output, but attenuated.
Our method is built on the theory. Defensive quantization not only quantizes feature maps into low-bit representations, but also controls the Lipschitz constant of the network, such that the noise is kept within a small magnitude for all hidden layers. In such case, we keep the noise small, as in the left zone of Figure~\ref{fig:dist_vs_epsilon}, quantization can reduce the perturbation error.
The produced model with our method enjoys better security and efficiency at the same time.

Experiments show that \methodnameverbose offers three unique advantages. First, DQ provides an effective way to boost the robustness of deep learning models while maintains the efficiency. Second, DQ is a generic building block of adversarial defense, which can be  combined with other adversarial defense techniques to advance state-of-the-art robustness. Third, our method makes quantization itself easier thanks to the constrained dynamic range.

\section{Background and Related Work}

\subsection{Model Quantization}

\begin{figure}[t]
\centering
\includegraphics[width=0.95\textwidth]{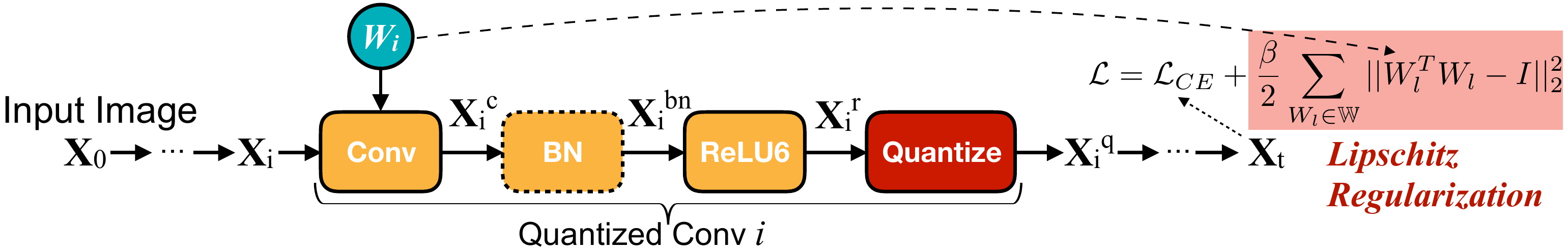}
\caption{Defensive quantization  with Lipschitz regularization.}
\label{fig:conv_blocks}
\end{figure}

Neural network quantization~\citep{han2015deep, rastegari2016xnor, zhou2016dorefa, courbariaux2016binarynet, zhu2016trained}) are widely adopted to enable efficient inference. By quantizing the network into low-bit representation, the inference of network requires less computation and less memory, while still achieves little accuracy degradation on clean images.
However, we find that the quantized models suffer from severe security issues --- they are more vulnerable against adversarial attacks compared to full-precision models, even when they have the same clean image accuracy.  
Adversarial perturbation is applied to the input image, thus it is most related to activation quantization~\citep{dhillon2018stochastic}. We carry out the rest of the paper using ReLU6 based activation quantization~\citep{howard2017mobilenets, sandler2018inverted}, as it is computationally efficient and is widely adopted by modern frameworks like TensorFlow~\citep{abadi2016tensorflow}. 
As illustrated in Figure~\ref{fig:conv_blocks}, a quantized convolutional network is composed of several quantized convolution block, each containing a serial of \emph{conv + BN + ReLU6 + linear quantize} operators.
As the quantization operator has 0 gradient almost everywhere, we followed common practice to use a STE~\citep{bengio2013estimating} function  $y = x + \text{stop\_gradient} [\text{Quantize}(x) - x]$ for gradient computation, which also eliminates the obfuscated gradient problem~\citep{athalye2018obfuscated}.

Our work bridges two domains:  model quantization and adversarial defense. Previous work~\citep{galloway2017attacking} claims binary quantized networks can improve the robustness against some attacks. However, the improvement is not substantial and they used randomized quantization, which is not practical for real deployment (need extra random number generators in hardware). It also causes one of the obfuscated gradient situations~\citep{athalye2018obfuscated}: stochastic gradients, leading to a false sense of security. \citet{rakin2018defend} tries to use quantization as an effective defense method. However, they employed Tanh-based quantization, which is not hardware friendly on fixed-point units due to the large overhead accessing look-up table. Even worse, according to our re-implementation, their method actually leads to severe gradient masking problem~\citep{papernot2016practical} during adversarial training, due to the nature of Tanh function (see~\ref{sec:app_tanh_gd} for detail). As a result, the actual robustness of this work under black-box attack has no improvement over full-precision model and is even worse. Therefore, there is no previous work that are conducted under real inference setting to study the quantized robustness for both black-box and white-box. Our work aim to raise people's awareness about the security of the actual deployed models.

\subsection{Adversarial Attacks \& Defenses}
Given an image $\mathbf{X}$, an adversarial attack method tries to find a small perturbation $\mathbf{\Delta}$ with constraint $||\mathbf{\Delta}|| \leq \epsilon$, such that the neural network gives different outputs for $\mathbf{X}$ and $\mathbf{X}^{adv} \triangleq \mathbf{X+\Delta}$. Here $\epsilon$ is a scalar to constrain the norm of the noise (\eg, $\epsilon=8$ is commonly used when we represent colors from 0-255), so that the perturbation is hardly visible to human. 
For this paper we choose to study attacks defined under  $||\cdot||_\infty$, where each element of the image can vary at most $\epsilon$ to form an adversary. We introduce several attack and defense methods used in our work in the following sections.

\subsubsection{Attack methods} 
\textbf{Random Perturbation (Random)} ~
Random perturbation attack adds a uniform sampled noise within $[-\epsilon, \epsilon]$ to the image, 
The method has no prior knowledge of the data and the network, thus is considered as the weakest attack method.

\textbf{Fast Gradient Sign Method (FGSM) \& R+FGSM} ~\citet{goodfellow6572explaining} proposed a fast method to calculate the adversarial noise by following the direction of the loss gradient $\nabla_{\mathbf{X}}L(\mathbf{X}, y)$, where $L(\mathbf{X}, y)$ is the loss function for training (\eg cross entropy loss). The adversarial samples are computed as:
\begin{equation}
    \mathbf{X}^{adv} = \mathbf{X} + \epsilon * \text{sign}(\nabla_{\mathbf{X}}L(\mathbf{X}, y))
\end{equation}
As FGSM is an one-step gradient-based method, it can suffer from sharp curvature near the data points, leading a false direction of ascent. Therefore, \citet{tramer2017ensemble} proposes to prepend FGSM by a random step to escape the non-smooth vicinity. The new method is called \textbf{R+FGSM}, defined as follows, for parameters $\epsilon$ and $\epsilon_1$ (where $\epsilon_1<\epsilon$):
\begin{equation}
    \mathbf{X}^{adv} = \mathbf{X^\prime} + (\epsilon - \epsilon_1) * \text{sign}(\nabla_{\mathbf{X}}L(\mathbf{X}, y)), \text{where}~ \mathbf{X}^\prime=\mathbf{X}+\epsilon_1 * \text{sign}(\mathcal{N}(\mathbf{0}^d, \mathbf{I}^d)).
\end{equation}
In our paper, we set $\epsilon_1 = \epsilon / 2$ following~\citet{tramer2017ensemble}. 

\textbf{Basic Iterative Method (BIM) \& Projected Gradient Descend (PGD)}~
\citet{kurakin2016adversarial} suggests a simple yet much stronger variant of FGSM by applying it multiple times iteratively with a small step size $\alpha$. The method is called BIM, defined as:
\begin{equation}
    \mathbf{X}^{adv}_0 = \mathbf{X},~ \mathbf{X}^{adv}_{n+1} = \text{clip}_{\mathbf{X}}^{\epsilon}\lbrace \mathbf{X}_n^{adv} + \alpha \text{sign}(\nabla_{\mathbf{X}}L(\textbf{X}_n^{adv}, y))\rbrace,
\end{equation}
where $\text{clip}_{\mathbf{X}}^{\epsilon}$ means clipping the result image to be within the $\epsilon$-ball of $\mathbf{X}$. In~\citep{madry2017towards}, the BIM is prepended by a random start as in R+FGSM method. The resulting attack is called PGD, which proves to be a general first-order attack. In our experiments we used PGD for comprehensive experiments as it proves to be one of the strongest attack. Unlike~\citet{madry2017towards} that uses a fixed $\epsilon$ and $\alpha$, we follow~\citet{kurakin2016adversarial, song2017pixeldefend} to use $\alpha=1$ and number of iterations of $\floor{\min(\epsilon + 4, 1.25\epsilon)}$, so that we can test the model's robustness under different strength of attacks.

\subsubsection{Defense methods} 
Current defense methods either preprocess the adversarial samples to denoise the perturbation~\citep{xu2017feature, song2017pixeldefend, liao2018defense} or making the network itself robust~\citep{warde201611, papernot2016distillation, madry2017towards, kurakin2016adversarial, goodfellow6572explaining, tramer2017ensemble}. Here we introduced several defense methods related to our experiments.

\textbf{Feature Squeezing} ~\citet{xu2017feature} proposes to detect adversarial images by squeezing the input image. Specifically, the image is processed with color depth bit reduction (5 bits for our experiments) and smoothed by a $2\times 2$ median filter. If the low resolution image is classified differently as the original image, then this image is detected as adversarial.

\textbf{Adversarial Training} Adversarial training~\citep{madry2017towards, kurakin2016adversarial, goodfellow6572explaining, tramer2017ensemble} is currently the strongest method for defense. By augmenting the training set with adversarial samples, the network learns to classify adversarial samples correctly. As adversarial FGSM can easily lead to gradient masking effect~\citep{papernot2016practical}, we study adversarial R+FGSM as in~\citep{tramer2017ensemble}. We also experimented with PGD training~\citep{madry2017towards}.

Experiments show that above defense methods can be combined with our DQ method to further improve the robustness. The robustness has been tested under the aforementioned attack methods.

\section{Conventional NN Quantization is Not Robust} \label{sec:inferior}

\begin{figure}[t]
\centering
\begin{subfigure}[b]{0.48\textwidth}
	\includegraphics[width=\textwidth]{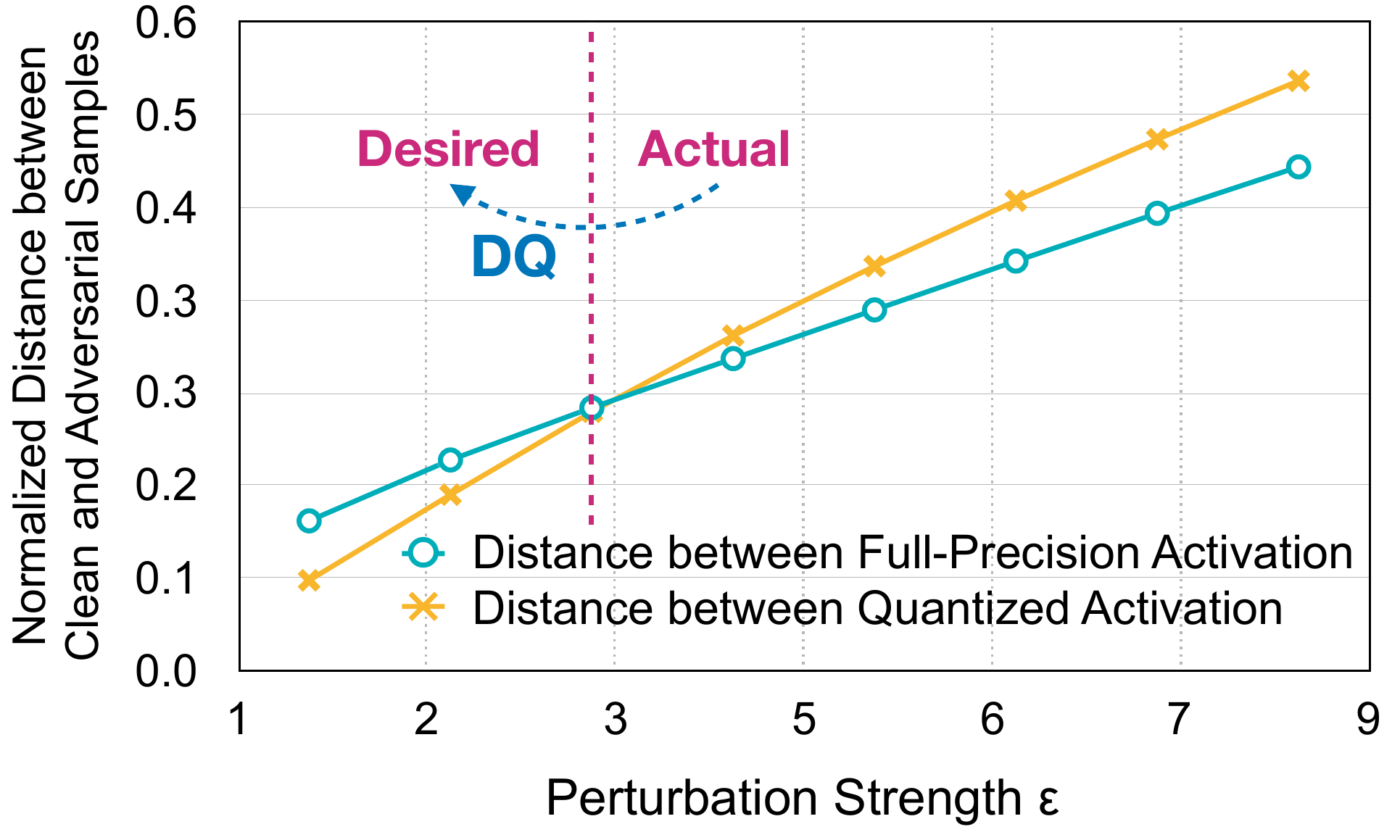}
	\caption{Noise increases with perturbation strength. Quantization makes the slope deeper.}
	\label{fig:dist_vs_epsilon}
\end{subfigure}
~~~
\begin{subfigure}[b]{0.48\textwidth}
	\includegraphics[width=\textwidth]{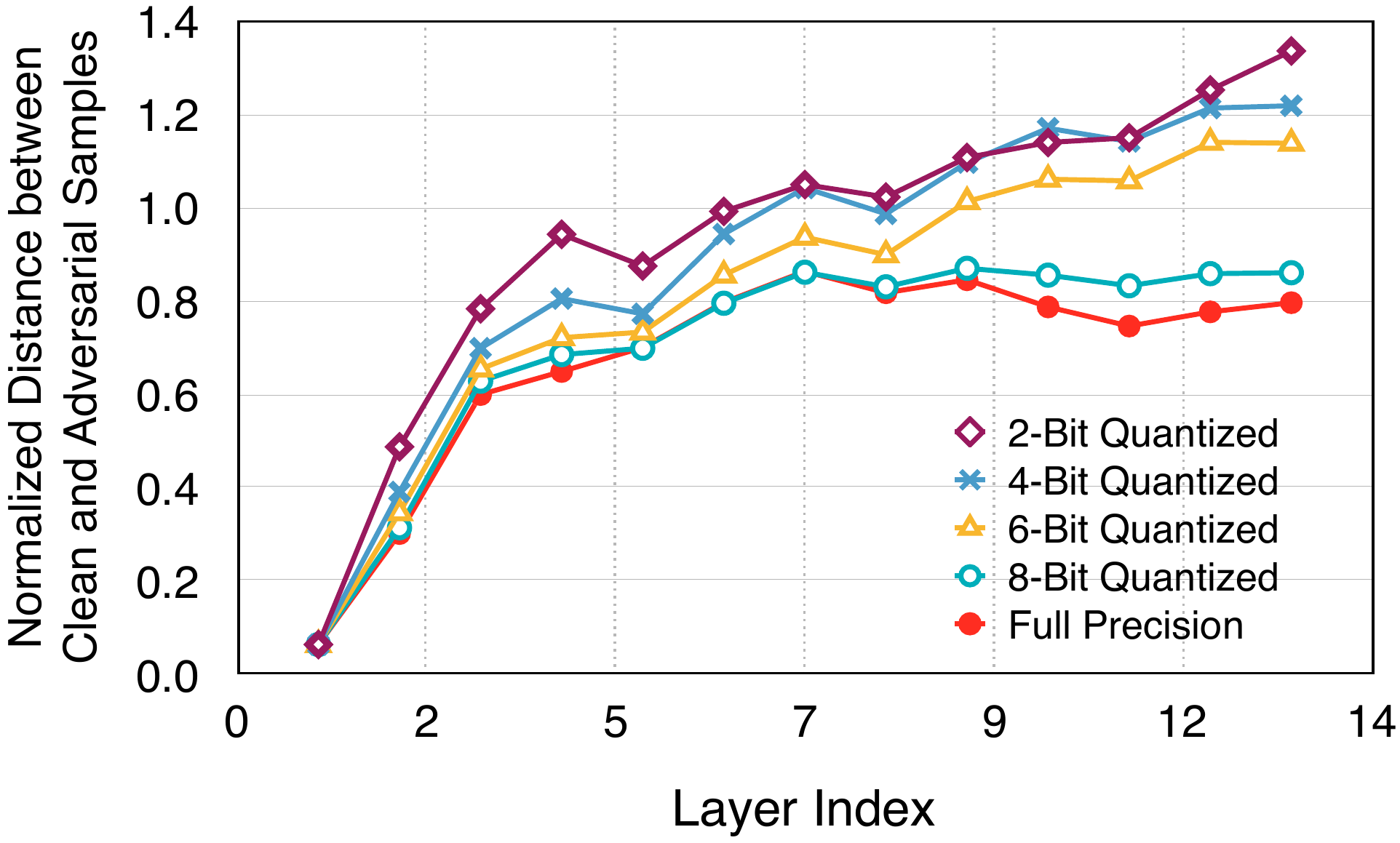}
	\caption{With conventional quantization, noise increases with layer index (the amplification effect).}
	\label{fig:vgg16_layer_diff}
\end{subfigure}
\caption{\textbf{(a)} Comparison of the noise introduced by adversarial-attack, with and without quantization. 
For small perturbation, quantization reduces the noise; for large perturbation, quantization magnifies the noise. \quad
\textbf{(b)} The noise amplification effect: the noise is amplified with layer index. Setup: conventional activation quantization for VGG-16, normalized difference of full-precision and low-precision activation.
}
\label{fig:explain}
\end{figure}

Conventional neural network quantization is more vulnerable to adversarial attacks. We experimented with VGG-16~\citep{simonyan2014very} and a Wide ResNet~\citep{zagoruyko2016wide} of depth 28 and width 10 on CIFAR-10~\citep{krizhevsky2009learning} dataset. 
We followed the training protocol as in~\citep{zagoruyko2016wide}.
Adversarial samples are generated with a FGSM~\citep{goodfellow6572explaining} attacker ($\epsilon=8$) on the entire test set. As in Figure~\ref{fig:acc_bit_curve}, the clean image accuracy doesn't significantly drop until the model is quantized to 4 bits (Figure~\ref{fig:clean_acc_bit_curve}). 
However, under adversarial attack, even with 5-bit quantization, the accuracy drastically decreased by $25.3\%$ and $9.2\%$ respectively. Although the full precision model's accuracy has dropped, the quantized model's accuracy dropped much harder, showing that the conventional quantization method is not robust. Clean image accuracy used to be the sole figure of merit to evaluate a quantized model. We show that even when the quantized model has no loss of performance on clean images, it can be much more easily fooled compared to full-precision ones, thus raising security concerns.

Input image quantization, \ie, color bit depth reduction is an effective defense method~\citep{xu2017feature}. Counter intuitively, it does not work when applied to hidden layers, and even make the robustness worse.
To understand the reason, we studied the effect of quantization \wrt different perturbation strength.
We first randomly sample 128 images $\mathbf{X}$ from the test set of CIFAR-10, and generate corresponding adversarial samples $\mathbf{X}^{adv}$. The samples are then fed to the trained Wide ResNet model. To mimic different strength of activation perturbation, we vary the $\epsilon$ from 1 to 8. We inspected the activation after the first convolutional layer $f_1$ (Conv + BN + ReLU6), denoted as $\mathbf{A}_1 = f_1(\mathbf{X})$ and $\mathbf{A}_1^{adv}=f_1(\mathbf{A}^{adv})$. To measure the influence of perturbation, we define a normalized distance between clean and perturbed activation as:
\begin{equation}
    D(\mathbf{A}, \mathbf{A}^{adv}) = ||\mathbf{A} - \mathbf{A}^{adv}||_2 / ||\mathbf{A}||_2
\end{equation}
We compare $D(\mathbf{A}, \mathbf{A}^{adv})$ and $D(\text{Quantize}(\mathbf{A}), \text{Quantize}(\mathbf{A}^{adv}))$, where $\text{Quantize}$ indicates uniform quantization with 3 bits. The results are shown in Figure~\ref{fig:dist_vs_epsilon}. We can see that only when $\epsilon$ is small, quantization helps to reduce the distance by removing small magnitude perturbations. The distance will be enlarged when $\epsilon$ is larger than 3. 

\begin{figure}[t]
\centering
\includegraphics[width=0.95\textwidth]{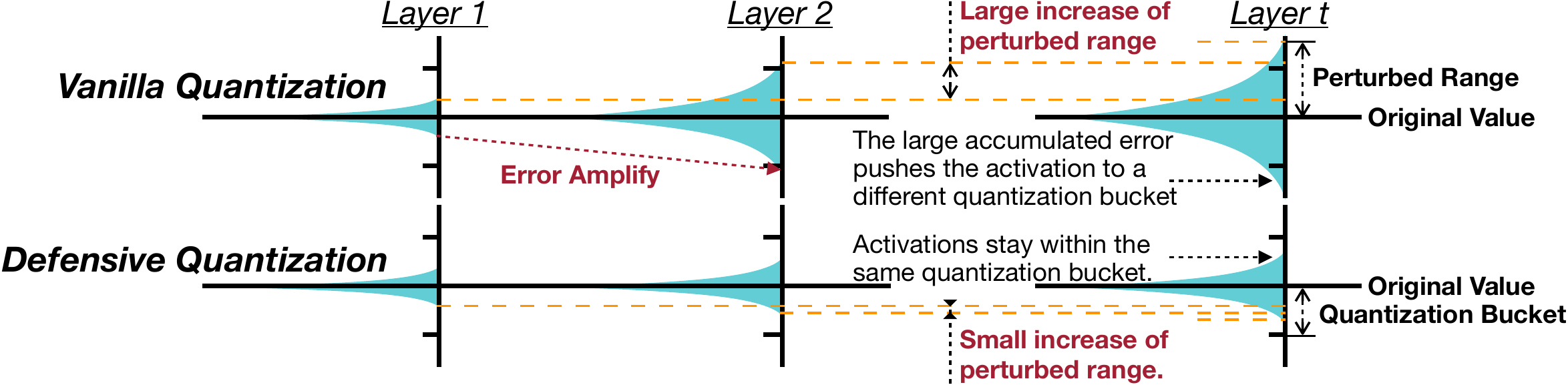}
\caption{The error amplification effect prevents activation quantization from defending adversarial attacks.}
\label{fig:noise_amplify}
\end{figure}

The above experiment explains the inferior robustness of the quantized model. We argue that such issue arises from the \textit{error amplification effect}~\citep{liao2018defense}, where the relative perturbed distance will be amplified when the adversarial samples are fed through the network. As illustrated in Figure~\ref{fig:noise_amplify}, the perturbation applied to the \emph{input image} has very small magnitude compared to the image itself ($\pm 8$ versus $0-255$), 
corresponding to the left zone of Figure~\ref{fig:dist_vs_epsilon} (desired), where quantization 
helps to denoise the perturbation.
Nevertheless, the difference in activation is amplified as the inference carries on. If the perturbation after amplification is large enough, 
the situation corresponds to the right zone (actual) of Figure~\ref{fig:dist_vs_epsilon}, where quantization further increases the normalized distance. 
Such phenomenon is also observed in the quantized VGG-16. We plot the normalized distance of each convolutional layer's input in Figure~\ref{fig:vgg16_layer_diff}. The fewer bits in the quantized model, the more severe the amplification effect.

\renewcommand \arraystretch{1.1}
\begin{table}[t]
\setlength{\tabcolsep}{6pt}
\caption{The clean and adversarial accuracy of Wide ResNet on CIFAR-10 test set. We compare the accuracy of full-precision and quantized models. With our DQ method, we not only eliminate the robustness gap between full-precision and quantized models, but also improve the robustness over full-precision ones. The accuracy gain from quantization compared to the full-precision model (Quantize Gain) has gradually been improved as $\beta$ increases. \textbf{Bold} and \underline{underline} numbers are the first and second highest accuracy at each row.}
\label{tab:fix_drop}
\small
\begin{center}
\begin{tabular}{cccccccccc}
\toprule[1.5pt]
\multirow{2}{*}{\textbf{Method}} & \multirow{2}{*}{\textbf{Acc.}} & \multirow{2}{*}{\begin{tabular}[c]{@{}c@{}} \textbf{Full}\\ \textbf{Prec.} \end{tabular}} & \multicolumn{5}{c}{\textbf{Quantize Bit}}  & \multirow{2}{*}{\begin{tabular}[c]{@{}c@{}} \textbf{Quantize}\\ \textbf{Gain} \end{tabular}} & \multirow{2}{*}{\begin{tabular}[c]{@{}c@{}} \textbf{Best}\\ \textbf{Acc.} \end{tabular}} \\ \cline{4-8} 
 &  &  & \textbf{1} & \textbf{2} & \textbf{3} & \textbf{4} & \textbf{5} \\ 
\toprule[1.5pt]
\multicolumn{1}{c}{\multirow{2}{*}{Vanilla}}     
& \multicolumn{1}{c}{clean}    & \multicolumn{1}{c}{94.8}   & \multicolumn{1}{c}{42.0}  & \multicolumn{1}{c}{84.9}  & \multicolumn{1}{c}{92.8}  & \multicolumn{1}{c}{93.9}  & \multicolumn{1}{c}{94.7} & \\ \cline{2-8} 
\multicolumn{1}{c}{} & ~adv.~ & \textbf{39.3} & 9.0 & 8.5 & 14.1 & 19.0 & \underline{30.2} & -9.1 & \textbf{39.3} \\ \toprule[1pt]
\multirow{2}{*}{DQ ($\beta=$3e-4)} & clean & 95.3 & 95.2 & 95.3 & 95.1 & 95.1  & 95.1  \\
 \cline{2-8}  & adv. & \underline{41.8} & \textbf{43.2}  & 41.1  & 40.1  &  39.8 & 39.3 & +1.4 &\textbf{43.2}\\ \toprule[1pt]
\multirow{2}{*}{DQ ($\beta=$6e-4)} & clean & 95.6 & 95.7 & 95.2 & 95.4 & 95.6 & 95.5 \\ 
\cline{2-8}  & adv. & 45.7 & \textbf{48.3} & \underline{47.9} &  43.8 & 43.9  & 44.6 & +2.6 & \textbf{48.3} \\ \toprule[1pt]
\multirow{2}{*}{DQ ($\beta=$1e-3)}  & clean  & 95.9  & 95.8  & 95.6 & 95.6  & 95.8 & 95.8 \\
\cline{2-8}  & adv.  & 49.1 & \underline{51.3}  & 50.4  & 51.3 & 49.8 & \textbf{51.8} & +2.7 & \textbf{51.8} \\ \bottomrule[1.5pt]
\end{tabular}
\end{center}
\end{table}

\section{\methodname} 
Given the robustness limitation of conventional quantization technique, we propose \methodnameverbose to defend the adversarial examples for quantized models. 
DQ suppresses the noise amplification effect, keeping the magnitude of the noise small, so that we can arrive at the left zone  (Figure~\ref{fig:dist_vs_epsilon}, desired) where quantization helps robustness instead of making it worse.

We control the neural network's \emph{Lipschitz constant}~\citep{szegedy2013intriguing, bartlett2017spectrally, cisse2017parseval} to suppress network's amplification effect. 
Lipschitz constant describes: when input changes, how much does the output change correspondingly. For a function $f: X\rightarrow Y$, if it satisfies
\begin{equation}
    D_Y(f(x_1), f(x_2)) \leq kD_X(x_1, x_2), ~~\forall x_1, x_2 \in X
\end{equation}
for a real-valued $k \geq 0$ and some metrics $D_X$ and $D_Y$, then we call $f$ Lipschitz continuous and $k$ is the known as the Lipschitz constant of $f$. If we consider a network $f$ with clean inputs $x_1=\mathbf{X}$ and corresponding adversarial inputs $x_2=\mathbf{X}_{adv}$, 
the error amplification effect can be controlled if we have a small Lipschitz constant $k$ (in optimal situation we can have $k \leq 1$). In such case, the error introduced by adversarial perturbation will not be amplified, but reduced. Specifically, we consider a feed-forward network composed of a serial of functions:
\begin{equation}
    f(x) = (\phi_l \circ \phi_{l-1} \circ ... \circ \phi_1)(x)
\end{equation}
where $\phi_i$ can be a linear layer, convolutional layer, pooling, activation functions, \etc. Denote the Lipschitz constant of a function $f$ as $Lip(f)$, then for the above network we have 
\begin{equation}
    Lip(f) \leq \prod_{i=1}^{L} Lip(\phi_i)
\end{equation}

As the Lipschitz constant of the network is the product of its individual layers' Lipschitz constants, $Lip(f)$ can grow exponentially if $Lip(\phi_i) > 1$. This is the common case for normal network training~\citep{cisse2017parseval}, and thus the perturbation will be amplified for such a network. Therefore, to keep the Lipschitz constant of the whole network small, we need to keep the Lipschitz constant of each layer $Lip(\phi_i) \leq 1$. We call a network with $Lip(\phi_i) \leq 1, \forall i=1, ..., L$ a non-expansive network.

We describe a regularization term to keep the Lipschitz constant small. Let us first consider linear layers with weight $\mathbf{W} \in \mathbb{R}^{c_{out}\times c_{in}}$ under $||\cdot||_2$ norm. The 
Lipschitz constant is by definition the spectral norm of $W$: $\rho(\mathbf{W})$, \ie, the maximum singular value of $\mathbf{W}$. 
Computing the singular values of each weight matrix is not computationally feasible during training. Luckily, if we can keep the weight matrix row orthogonal, the singular values are by nature equal to 1, which meets our non-expansive requirements. Therefore we transform the problem of keeping $\rho(\mathbf{W}) \leq 1$ into keeping $\mathbf{W}^T\mathbf{W} \approx \mathbf{I}$, where $\mathbf{I}$ is the identity matrix. Naturally, we introduce a regularization term $ || \mathbf{W}^T\mathbf{W} - \mathbf{I} ||$,
where $I$ is the identity matrix. Following~\citep{cisse2017parseval}, for convolutional layers with weight $\mathbf{W} \in \mathbb{R}^{c_{out} \times c_{in} \times k \times k}$, we can view it as a two-dimension matrix of shape $\mathbf{W} \in \mathbb{R}^{c_{out} \times (c_{in} k k)}$ and apply the same regularization. 
The final optimization objective is:
\begin{equation}
    \mathcal{L} = \mathcal{L}_{CE} + \frac{\beta}{2} \sum_{\mathbf{W}_l\in \mathbb{W}} || \mathbf{W}_l ^T\mathbf{W}_l - \mathbf{I} ||^2
\end{equation}

where $\mathcal{L}_{CE}$ is the original cross entropy loss and $\mathbb{W}$ denotes all the weight matrices of the neural network. $\beta$ is the weighting to adjust the relative importance. The above discussion is based on simple feed forward networks. For ResNets in our experiments, we also follow~\citet{cisse2017parseval} to modify the aggregation layer as a convex combination of their inputs, where the 2 coefficients are updated using specific projection algorithm (see \citep{cisse2017parseval} for details).

Our \methodname is illustrated in Figure~\ref{fig:conv_blocks}. The key part is the regularization term, which suppress the noise amplification effect by regularizing the Lipschitz constant. As a result, the perturbation at each layer is kept within a certain range, the adversarial noise won't propagate. 
Our method not only fixes the drop of robustness induced by quantization, but also takes quantization as a defense method to further increase the robustness. Therefore it's named \methodname. 
\section{Experiments}

Our experiments demonstrate the following advantages of \methodname. First, DQ can retain the robustness of a model when quantized with low-bit. Second, DQ is a general and effective defense method under various scenarios, thus can be combined with other defensive techniques to further advance state-of-the-art robustness. Third, as a by-product, DQ can also improve the accuracy of training quantized models on clean images without attacks, since it limits the dynamic range. 
\subsection{Fixing Robustness Drop} \label{sec:fix_robust_drop}

\textbf{Setup:} We conduct experiments with Wide ResNet~\citep{zagoruyko2016wide} of $28\times 10$ on the CIFAR-10 dataset~\citep{krizhevsky2009learning} using ReLU6 based activation quantization, with number of bits ranging from 1 to 5. All the models are trained following~\citep{zagoruyko2016wide} with momentum SGD for 200 epochs. The adversarial samples are generated using FGSM attacker with $\epsilon=8$.
 
\textbf{Result:} The results are presented in Table~\ref{tab:fix_drop}. For vanilla models, though the adversarial robustness increases with the number of bits, \ie, the models closer to full-precision one has better robustness, the best quantized model still has inferior robustness by $-9.1\%$. While with our \methodname, the quantized models have better robustness than full-precision counterparts. The robustness is better when the number of bits are small, since it can de-noise larger adversarial perturbations. We also find that the robustness is generally increasing as $\beta$ gets larger, since the regularization of Lipschitz constant itself keeps the noise smaller at later layers. At the same time, the quantized models consistently achieve better robustness. The robustness of quantized model also increases with $\beta$. We conduct a detailed analysis of the effect of $\beta$ in Section~\ref{sec:app_beta_study}.
The conclusion is: (1) conventional quantized models are less robust. (2) Lipschitz regularization makes the model robust. (3) Lipschitz regularization + quantization makes model even more robust.

\begin{figure}[t]
\centering
\begin{subfigure}[b]{0.48\textwidth}
	\includegraphics[width=\textwidth]{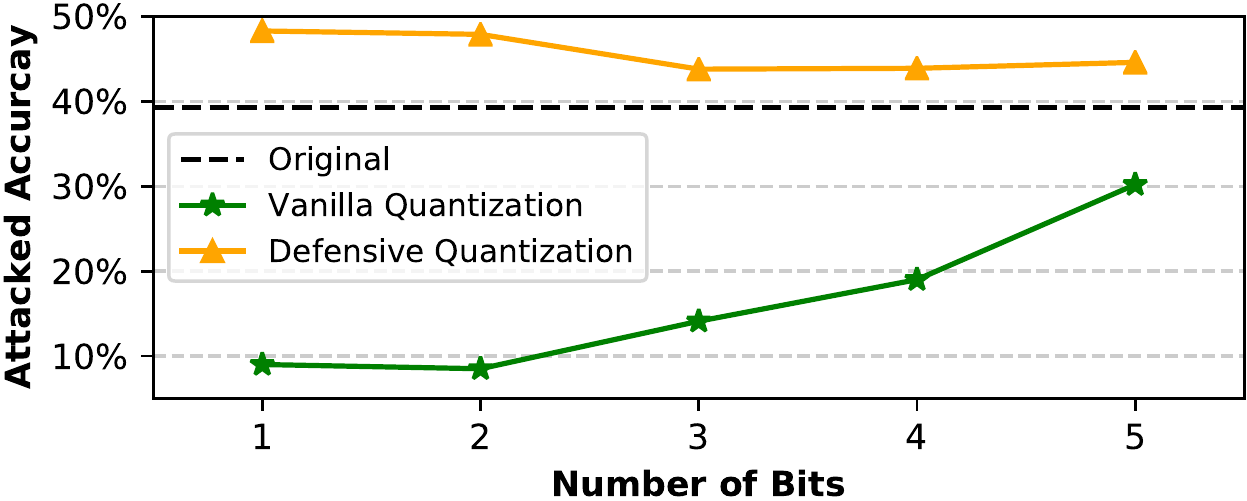}
	\caption{White-Box Robustness ($\epsilon=8$)}
	\label{fig:white_box_eps8}
\end{subfigure}
~
\begin{subfigure}[b]{0.48\textwidth}
	\includegraphics[width=\textwidth]{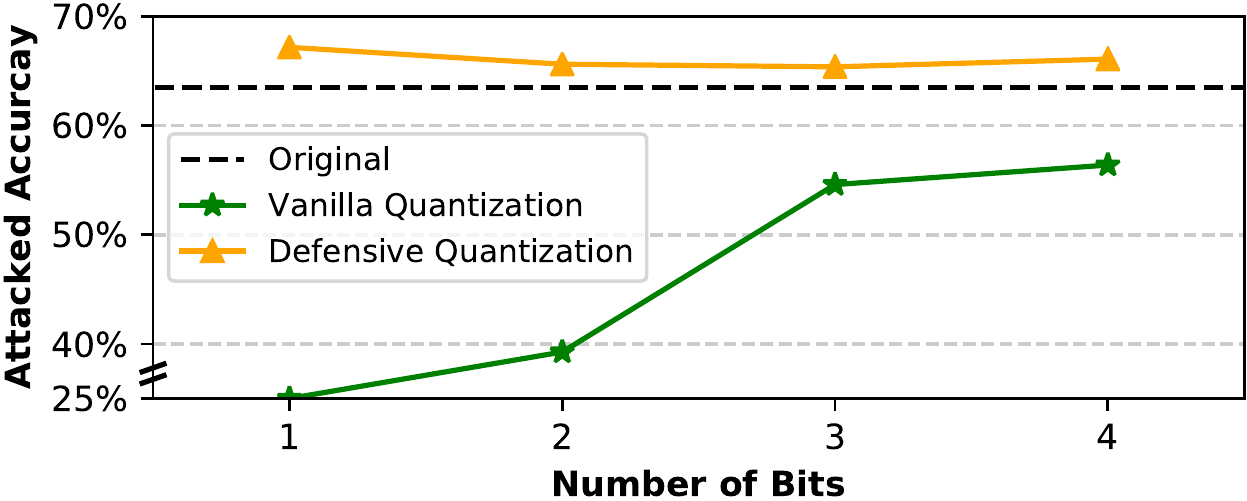}
	\caption{Black-Box Robustness ($\epsilon=8$)}
	\label{fig:black_box_eps8}
\end{subfigure}
\caption{
The white-box and black-box robustness are consistent: vanilla quantization leads to significant robustness drop, while DQ can bridge the gap and improve the robustness, especially with lower bits (bit=1).
Setup: white-box and black-box robustness of Wide ResNet with vanilla quantization and defensive quantization. 
}
\label{fig:black_box}
\end{figure}

As shown in~\citep{athalye2018obfuscated}, many of the defense methods actually lead to obfuscated gradient, providing a false sense of security. Therefore it is important to check the model's robustness under black-box attack. We separately trained a substitute VGG-16 model on the same dataset to generate adversarial samples, as it was proved to have the best transferability~\citep{su2018robustness}. 
The results are presented in Figure~\ref{fig:black_box}. Trends of white-box and black-box attack are consistent. Vanilla quantization leads to inferior black-box robustness,
while with our method
can further improve the models' robustness.
As the robustness gain is consistent for both white-box and black-box setting, our method does not suffer from gradient masking.

\subsection{Defend with \methodname}

In this section, we show that we can combine \methodname with other defense techniques to achieve 
state-of-the-art robustness.

\textbf{Setup:}  We conducted experiments on the Street View House Number dataset (SVHN)~\citep{netzer2011reading} and CIFAR-10 dataset~\citep{krizhevsky2009learning}.
Since adversarial training is time consuming, we only use the official training set for experiments.
CIFAR-10 is another widely used dataset containing 50,000 training samples and 10,000 testing samples of size $32\times 32$. For both datasets, we divide the pixel values by 255 as a pre-processing step.

Following~\citep{athalye2018obfuscated, cisse2017parseval, madry2017towards}, we used Wide ResNet~\citep{zagoruyko2016wide} models in our experiments as it is considered as the standard model on the dataset. We used depth 28 and widening factor 10 for CIFAR-10, and depth 16 and widening factor 4 for SVHN. We followed the training protocol in~\citep{zagoruyko2016wide} that uses a SGD optimizer with momentum=0.9. For CIFAR-10, the model is trained for 200 epochs with initial learning rate 0.1, decayed by a factor of 0.2 at 60, 120 and 160 epochs. For SVHN dataset, the model is trained for 160 epochs, with initial learning rate 0.01, decayed by 0.1 at 80 and 120 epochs. 
For DQ, we used bit=1 and $\beta=\text{2e-3}$ as it offers the best robustness (see Section~\ref{sec:app_beta_study}).

We combine DQ with other defense methods to further boost the robustness. For Feature Squeezing~\citep{xu2017feature}, we used 5 bit for image color reduction, followed by a $2\times 2$ median filter. As adversarial FGSM training leads to gradient masking issue~\citep{tramer2017ensemble} (see~\ref{sec:app_fgsm_gd} for our experiment), we used the variant adversarial R+FGSM training. To avoid over-fitting into certain $\epsilon$, we randomly sample $\epsilon$ using $\epsilon^\prime \sim \mathcal{N}(0, \delta), \epsilon = \text{clip}^{[0, 2\delta]}(\text{abs}(\epsilon^\prime))$. Specifically we used $\delta=8$ to cover $\epsilon$ from 0-16. During test time, the $\epsilon$ is set to a fixed value (2/8/16). We also conducted adversarial PGD training. Following~\citep{kurakin2016adversarial, song2017pixeldefend}, during training we sample random $\epsilon$ as in R+FGSM setting, and generate adversarial samples using step size 1 and number of iterations $\floor{\min(\epsilon + 4, 1.25\epsilon)}$. 

\textbf{Result:} The results are presented in Table~\ref{tab:svhn_exps} and Table~\ref{tab:cifar10_exps}, where \textbf{(B)} indicates black-box attack with a seperately trained VGG-16 model. The bold number indicates the best result in its column.
We observe that for all normal training, feature squeezing and adversarial training settings, our DQ method can further improve the model's robustness. Among all the defenses, adversarial training provides the best performance against various attacks, epecially adversarial R+FGSM training. While white box PGD attack is generally the strongest attack in our experiments. Our DQ method also consistently improves the black-box robustness and there is no sign of gradient masking. Thus DQ proves to be an effective defense for various white-box and black-box attacks.

\renewcommand \arraystretch{1.3}
\begin{table}[t]
\setlength{\tabcolsep}{3pt}
\caption{SVHN experiments tested with $\epsilon=2/8/16$. \textbf{(B)} indicates black-box attack. }
\label{tab:svhn_exps}
\small
\begin{center}
\begin{tabular}{cccccccc}
\toprule[1.5pt]
\textbf{Training Technique}
& \textbf{Clean} & \textbf{Random}  & \textbf{FGSM} & \textbf{R+FGSM} & \textbf{PGD} & 
\textbf{FGSM(B)} & \textbf{PGD(B)}  \\ 
\toprule[1.5pt]
Normal & \textbf{96.8} & \textbf{97}/\textbf{97}/\textbf{96} & 74/42/26 & 86/55/35 & 77/05/00 & 90/62/40 & 92/58/30 \\ 
Normal + \textbf{\textit{DQ}} & 96.7 & 96/96/\textbf{96} & 77/45/31 & 87/59/40 & 79/10/00 & 90/63/42 & 92/60/31 \\  \hline
Feature Squeezing & 96.3 & 96/96/95 & 69/34/20 & 84/48/28 & 72/03/00 & 89/61/38 & 91/56/27  \\ 
Feature Squeezing + \textbf{\textit{DQ}} & 96.2 & 96/96/\textbf{96} & 75/42/28 & 86/56/36 & 77/06/00 & 90/63/40 & 92/59/30  \\  \hline
Adversarial R+FGSM & 96.6 & \textbf{97}/96/\textbf{96} & 84/53/38 & 91/70/57 & 87/30/06 & 93/74/54 & 94/83/72 \\ 
Adversarial R+FGSM + \textbf{\textit{DQ}}& 96.6 & \textbf{97}/96/95 & \textbf{88}/\textbf{59}/\textbf{40} & \textbf{93}/\textbf{77}/\textbf{61} & \textbf{91}/\textbf{47}/\textbf{17} & \textbf{94}/\textbf{80}/\textbf{61} & \textbf{95}/\textbf{89}/\textbf{84} \\
\bottomrule[1.5pt]
\end{tabular}
\end{center}
\end{table}

\renewcommand \arraystretch{1.3}
\begin{table}[t]
\setlength{\tabcolsep}{3pt}
\caption{CIFAR-10 Experiments tested with $\epsilon=2/8/16$.  \textbf{(B)} indicates black-box attack.}
\label{tab:cifar10_exps}
\small
\begin{center}
\begin{tabular}{cccccccc}
\toprule[1.5pt]
 \begin{tabular}[c]{@{}c@{}} \textbf{Training Technique} \end{tabular} 
& \textbf{Clean} & \textbf{Random}  & \textbf{FGSM} & \textbf{R+FGSM} & \textbf{PGD} & \textbf{FGSM(B)} & \textbf{PGD(B)}   \\ 
\toprule[1.5pt]
 Normal & 94.8 & 95/91/77 & 59/39/29 & 72/37/17 & 56/01/00 & 88/63/36 & 90/64/42 \\ %
 Normal + \textbf{\textit{DQ}} & \textbf{95.9} & \textbf{96}/\textbf{94}/84 & 68/53/42 & 77/50/30 & 62/04/00 & 84/59/40 & 87/50/27 \\  \hline  %
 Feature Squeezing & 94.1 & 94/92/81 & 61/35/27 & 76/40/21 & 64/02/00 & 87/62/30 & 89/70/42 \\ %
 Feature Squeezing + \textbf{\textit{DQ}} & 94.9 & 95/93/82 & 66/48/33 & 77/51/29 & 66/06/01 & 87/62/29 & 90/70/43 \\   \hline  %
 Adversarial R+FGSM & 91.6 & 92/91/91 & 81/52/38 & 87/69/48 & 84/43/11 & 92/89/85 & 92/91/89\\ %
 Adversarial R+FGSM + \textbf{\textit{DQ}}& 94.0 & 94/93/\textbf{93} & \textbf{85}/\textbf{63}/\textbf{51} & \textbf{90}/\textbf{74}/\textbf{58} & \textbf{87}/\textbf{50}/\textbf{22} & \textbf{93}/\textbf{91}/\textbf{87} & \textbf{94}/\textbf{92}/\textbf{92} \\  \hline %
 Adversarial PGD & 86.6 & 86/86/86 & 74/46/31 & 79/63/46 & 76/44/20 & 84/83/81 & 84/83/83 \\ %
 Adversarial PGD + \textbf{\textit{DQ}}& 87.5 & 87/87/87 & 79/53/36 & 83/69/52 & 81/\textbf{50}/\textbf{22} & 87/86/82 & 87/86/86 \\ %
\bottomrule[1.5pt]
\end{tabular}
\end{center}
\end{table}

\renewcommand \arraystretch{1}
\begin{table}[h!]
\setlength{\tabcolsep}{7pt}
\caption{DQ method improves the training of normal quantized models by limiting the dynamic range of activation. With conventional quantization, ReLU1 suffers from inferior performance than ReLU6. While with DQ, the gap is fixed, ReLU1 and ReLU6 quantized models achieve similar accuracy.}
\label{tab:improve_training}
\small
\begin{center}
\begin{tabular}{cccc}
\toprule[1.5pt]
& \textbf{ReLU1} & \textbf{ReLU6}  & \textbf{Difference} \\ 
\midrule[1.5pt]
Vanilla Quantization & 93.49\% & 94.41\% & -0.92\% \\ %
Defensive Quantization & 95.14\% & 95.09\% & 0.05\% \\
\bottomrule[1.5pt]
\end{tabular}
\end{center}
\end{table}

\subsection{Improve the Training of Quantized Models}

As a by-product of our method, \methodname can even improve the accuracy of quantized models on clean images without attack, making it a beneficial drop-in substitute for normal quantization procedures.
Due to conventional quantization method's amplification effect, the distribution of activation can step over the truncation boundary (0-6 for ReLU6, 0-1 for ReLU1), which makes the optimization difficult. DQ explicitly add a regularization to shrink the dynamic range of activation, so that it is fitted within the truncation range. To demonstrate our hypothesis, we experimented with ResNet and CIFAR-10. We quantized the activation with 4-bit (because NVIDIA recently introduced INT4 in Turing architecture) using ReLU6 and ReLU1 respectively. Vanilla quantization and DQ training are conducted for comparison. As shown in Table~\ref{tab:improve_training}, with vanilla quantization, ReLU1 model has around $1\%$ worse accuracy than ReLU6 model, although they are mathematically equal if we multiply the previous BN scaling by 1/6 and next convolution weight by 6. It demonstrates that improper truncation function and range will lead to training difficulty. While with DQ training, both model has improved accuracy compared to vanilla quantization, and the gap between ReLU1 model and ReLU6 model is filled, making quantization easier regardless of truncation range.
\section{Conclusion}
In this work, we aim to raise people's awareness about the security of the quantized neural networks, which is widely deployed in GPU/TPU/FPGAs, and pave a possible direction to bridge two important areas in deep learning: efficiency and robustness. 
We connect these two domains by designing a novel \methodnameverbose module to defend adversarial attacks while maintain the efficiency. Experimental results on two datasets validate that the new quantization method can make the deep learning models be safely deployed on mobile devices.

\subsubsection*{Acknowledgments}
We thank the support from MIT Quest for Intelligence, MIT-IBM Watson AI Lab, MIT-SenseTime Alliance, Xilinx, Samsung and AWS Machine Learning Research Awards.

\bibliography{egbib}
\bibliographystyle{iclr2019_conference}

\appendix
\section{Gradient Masking of Other Methods}
\subsection{Tanh-based Activation Quantization Leads to Fake Security} \label{sec:app_tanh_gd}

\citet{rakin2018defend} tried to use activation quantization as a defense method against adversarial samples. According to their experiments, simply Tanh based activation quantization can greatly improve the model's robustness under PGD adversarial training setting. We reproduced their experiments by training a ResNet-18 model with PGD adversarial training on CIFAR-10 following~\citet{madry2017towards}. To make a comparison, we also trained a full precision model and a ReLU6 quantized model following same setting. All the quantized models use bit=2. We tested the trained model using PGD attack by~\citet{madry2017towards} under both white-box and black-box setting. We use a VGG-16 model separately trained with PGD adversarial training as the black-box attacker. The results are provided in Table~\ref{tab:app_tanh_gm}. 

\renewcommand \arraystretch{1.2}
\begin{table}[h]
\setlength{\tabcolsep}{7pt}
\caption{Tanh-based quantization improves white-box robustness, while suffers from inferior black-box robustness compared to full-precision model. The white-box accuracy is even higher than black-box, suggesting gradient masking.
Setup: white-box and black-box robustness of PGD adversarially trained ResNet-18 on CIFAR-10. Quantization uses 2 bits.}
\label{tab:app_tanh_gm}
\small
\begin{center}
\begin{tabular}{cccc}
\toprule [1.5pt]
& \textbf{Clean Acc.} & \textbf{White-Box Acc.}  & \textbf{Black-Box Acc.} \\ 
\midrule [1.5pt]
Full-Precision & 83.47\% & 50.51\% & 65.58\% \\
Tanh Quantized & 81.60\% & 71.03\% & 61.57\% \\
ReLU6 Quantized & 78.97\% & 48.90\% & 59.44\% \\
\bottomrule [1.5pt]
\end{tabular}
\end{center}
\end{table}

Our white-box result is consistent with~\citep{rakin2018defend}, where Tanh based quantizaion with PGD training gives much higher white-box accuracy compared to the full precision model. However, we can see that the black-box robustness decreases. Worse still, the black-box attack successful rate is even lower than white-box, which is abnormal since black-box attack is generally weaker than white-box. This phenomenon indicates severe gradient masking problem~\citep{papernot2016practical, athalye2018obfuscated}, which gives a false sense of security. As a comparison, we also trained a ReLU6 quantized model. With ReLU6 quantization, there is no sign of gradient masking, nor improvement in robustness, indicating that gradient masking problem majorly comes from Tanh activation function. Actually, ReLU6 quantized model has slightly worse robustness.

Therefore we conclude that simple quantization cannot help to improve robustness. Instead, it leads to inferior robustness.

\subsection{Adversarial FGSM Training Causes Gradient Masking} \label{sec:app_fgsm_gd}
Here we demonstrate that FGSM adversarial training leads to significant gradient masking problem, while R+FGSM fixes such issues. We trained a Wide ResNet using adversarial FGSM and adversarial R+FGSM respectively. Then the model is tested using FGSM with $\epsilon=8$ under white-box and black-box setting (VGG-16 as substitute). The results are shown in Table 

\renewcommand \arraystretch{1.2}
\begin{table}[h]
\setlength{\tabcolsep}{7pt}
\caption{White-box and black-box robustness of FGSM/R+FGSM adversarially trained Wide ResNet on CIFAR-10.}
\label{tab:app_fgsm_gm}
\small
\begin{center}
\begin{tabular}{cccc}
\toprule [1.5pt]
& \textbf{Clean Acc.} & \textbf{White-Box Acc.}  & \textbf{Black-Box Acc.} \\ 
\midrule [1.5pt]
Adversarial FGSM & 94.55\% & 94.18\% & 51.65\% \\
Adversarial R+FGSM & 91.61\% & 51.65\% & 87.39\% \\
\bottomrule[1.5pt]
\end{tabular}
\end{center}
\end{table}

We can clearly see the same gradient masking effect. For adversarial FGSM training, it gives much higher white box robustness than R+FGSM adversarial training, while the black-box robustness is much worse. To avoid gradient masking, we thus use R+FGSM adversarial training instead.

\section{Hyper-Parameters: $\beta$ Study } \label{sec:app_beta_study}

\begin{figure}[h]
\centering
\includegraphics[width=0.7\textwidth]{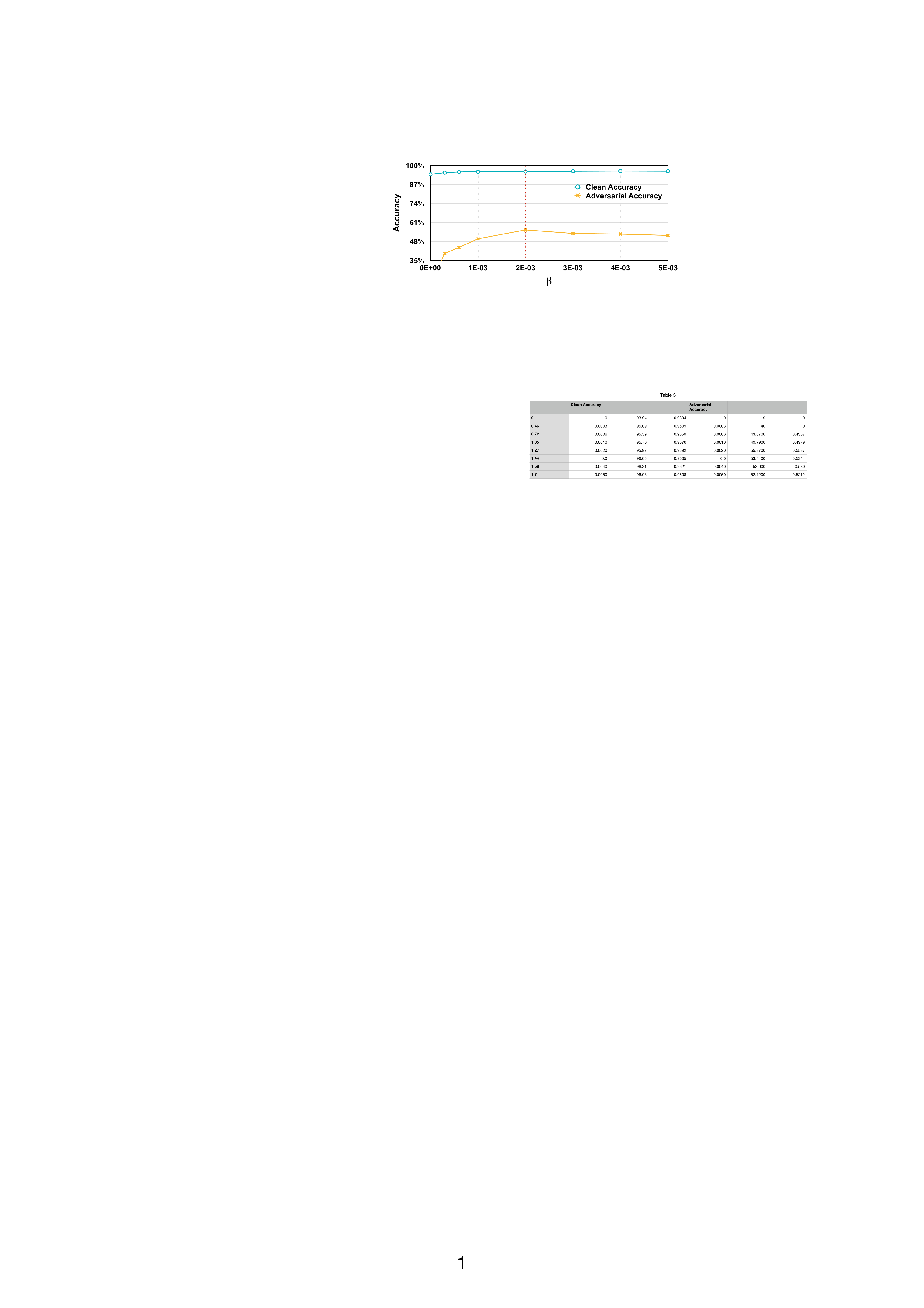}
\caption{Clean and adversarial accuracy of 4-bit quantized Wide ResNet \wrt different $\beta$. }
\label{fig:beta_study}
\end{figure}
As observed in Section~\ref{sec:fix_robust_drop}, the adversarial robustness of the model generally increases as $\beta$ gets larger. Here we aim to find what is the optimal $\beta$ for our experiment. We took 4-bit quantized Wide ResNet for example. As shown in Figure~\ref{fig:beta_study}, the adversarial accuracy first gets larger as $\beta$ increases, and slowly goes down afterwards, reaching peak performance at $\beta=0.002$. The clean accuracy is more stable. At the first stage, as regularization gets stronger, the effect of noise amplify is suppressed more. While for the second stage, the training suffers from a too strong regularization. Therefore, we used a $\beta=0.002$ for our experiments unless specified.

\newpage
\section{Visualize Samples and Predictions}

We visualize some of the clean and adversarial samples from CIFAR-10 test set, and the corresponding prediction for full precision model (\textbf{FP}), vanilla 4-bit quantized model (\textbf{VQ}) and our 4-bit Defensive Quantization model (\textbf{DQ}), as 4-bit quantization is now now supported by Tensor Cores with NVIDIA's Turing Architecture~\citep{nv_turing}. Predicted class and probability is provided. Compared to FP and DQ, VQ has worse robustness by misclassifying more adversarial samples (sample 1, 2, 4, 6). Our DQ model enjoys better robustness than FP model in two aspects: 1. our DQ model succeeds to defend some of the attacks when FP failed (sample 5, 7); 2. our DQ model has a better confidence for true label compared to FP model when both succeeds to defend (sample 1, 4, 6). Even when all models fail to defend, our DQ model has the lowest confidence for the misclassified class (sample 3, 8).

\begin{figure}[h]
\centering
\includegraphics[width=0.85\textwidth]{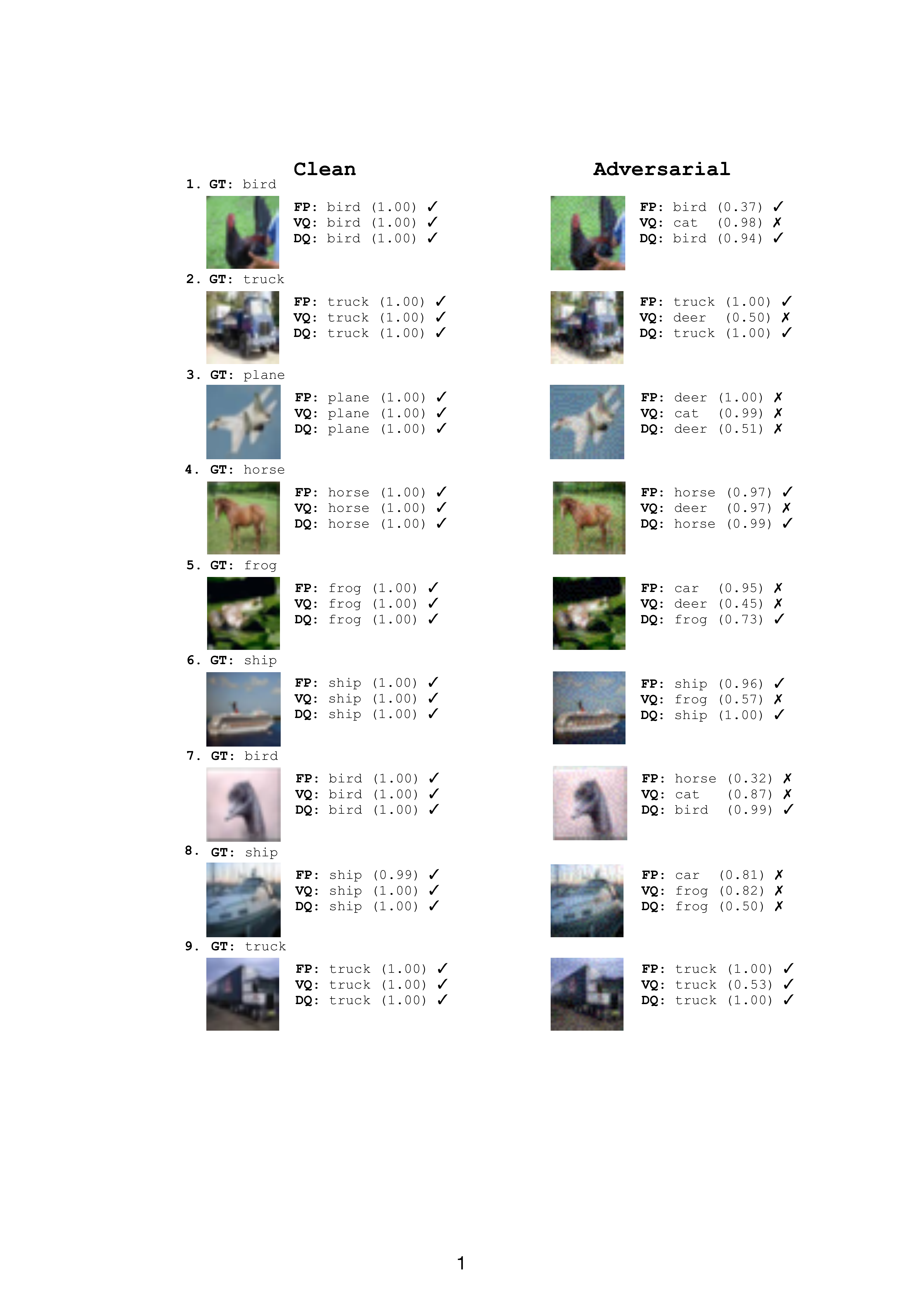}
\caption{Visualization of adversarial samples and the corresponding predictions (label and probability). \textbf{FP}: full-precision model; \textbf{VQ}: vanilla 4-bit quantized model; \textbf{DQ}: our 4-bit Defensive Quantization model}
\label{fig:app_svhn_sample}
\end{figure}

\end{document}